\documentclass[]{interact}

\usepackage{epstopdf}% To incorporate .eps illustrations using PDFLaTeX, etc.
\usepackage[caption=false]{subfig}% Support for small, `sub' figures and tables

\usepackage[longnamesfirst,sort]{natbib}% Citation support using natbib.sty
\bibpunct[, ]{(}{)}{;}{a}{,}{,}% Citation support using natbib.sty
\setcitestyle{aysep={}} 
\renewcommand\bibfont{\fontsize{10}{12}\selectfont}% To set the list of references in 10 point font using natbib.sty

\usepackage{colortbl}
\definecolor{Green}{rgb}{0.6,1,0.6}
\usepackage{url}
\usepackage{comment}
\usepackage{wrapfig}
\theoremstyle{plain}% Theorem-like structures provided by amsthm.sty

\theoremstyle{definition}

\theoremstyle{remark}

\begin{document}

%\articletype{ARTICLE TEMPLATE}% Specify the article type or omit as appropriate

\title{Learning Bidirectional Action-Language Translation with Limited Supervision and Testing with Incongruent Input}

\author{
\name{Ozan Özdemir\textsuperscript{a}\thanks{CONTACT Ozan Özdemir. Email: ozan.oezdemir@uni-hamburg.de}, Matthias Kerzel\textsuperscript{a}, Cornelius Weber\textsuperscript{a}, Jae Hee Lee\textsuperscript{a}, Muhammad Burhan Hafez\textsuperscript{a}, Patrick Bruns\textsuperscript{b}, Stefan Wermter\textsuperscript{a}}
\affil{\textsuperscript{a}Knowledge Technology, Department of Informatics, University of Hamburg, Vogt-Koelln-Str. 30, 22527 Hamburg, Germany}
\affil{\textsuperscript{b}Biological Psychology and Neuropsychology, University of Hamburg, Von-Melle-Park 11, 20146 Hamburg, Germany}
}

%\author{
%\name{A.~N. Author\textsuperscript{a}\thanks{CONTACT A.~N. Author. Email: %latex.helpdesk@tandf.co.uk} and John Smith\textsuperscript{b}}
%\affil{\textsuperscript{a}Taylor \& Francis, 4 Park Square, Milton Park, %Abingdon, UK; \textsuperscript{b}Institut f\"{u}r Informatik, Albert-Ludwigs-%Universit\"{a}t, Freiburg, Germany}
%}

\maketitle

\begin{abstract}
Human infant learning happens during exploration of the environment, by interaction with objects, and by listening to and repeating utterances casually, which is analogous to unsupervised learning. Only occasionally, a learning infant would receive a matching verbal description of an action it is committing, which is similar to supervised learning. Such a learning mechanism can be mimicked with deep learning. We model this weakly supervised learning paradigm using our Paired Gated Autoencoders (PGAE) model, which combines an action and a language autoencoder. After observing a performance drop when reducing the proportion of supervised training, we introduce the Paired Transformed Autoencoders (PTAE) model, using Transformer-based crossmodal attention. PTAE achieves significantly higher accuracy in language-to-action and action-to-language translations, particularly in realistic but difficult cases when only few supervised training samples are available. We also test whether the trained model behaves realistically with conflicting multimodal input. In accordance with the concept of incongruence in psychology, conflict deteriorates the model output. Conflicting action input has a more severe impact than conflicting language input, and more conflicting features lead to larger interference. PTAE can be trained on mostly unlabelled data where labeled data is scarce, and it behaves plausibly when tested with incongruent input.
\end{abstract}

\begin{keywords}
Unsupervised learning; weak supervision; autoencoders; object manipulation; robot action; language grounding; Transformers; bidirectional translation
\end{keywords}

\section{Introduction}

Embodiment, i.e., action-taking in the environment, is considered essential for language learning (Bisk et al. \citeyear{bisk-etal-2020-experience}). Recently, language grounding with robotic object manipulation has received considerable attention from the research community. Most approaches proposed in this domain cover robotic action execution based on linguistic input (Hatori et al. \citeyear{hatori2018interactively}; Shridhar, Mittal, and Hsu \citeyear{shridhar2018interactive}; Shao et al. \citeyear{shao2020concept2robot}; Lynch and Sermanet \citeyear{lynch2021language}), i.e., language-to-action translation. Others cover language production based on the actions done on objects (Heinrich et al. \citeyear{heinrich2020}; Eisermann et al. \citeyear{ELWW21}), i.e., action-to-language translation. However, only few approaches (Ogata et al. \citeyear{ogata2007parametricbias}; Yamada et al. \citeyear{yamada2018paired}; Antunes et al. \citeyear{antunes2019multitimescale}; Abramson et al. \citeyear{abramson2020imitating}; Özdemir, Kerzel, and
Wermter \citeyear{Oezdemir_2021_ICDL}) handle both directions by being able to not just execute actions according to given instructions but also to describe those actions, i.e., bidirectional translation.

Moreover, as infants learn, the actions that they are performing are not permanently labeled by matching words from their caretakers, hence, supervised learning with labels must be considered rare. Instead, infants rather explore the objects around them and listen to utterances, which may not frequently relate to their actions, hence, unsupervised learning without matching labels is abundant. Nevertheless, most language grounding approaches do not make use of unsupervised learning except those that use some unsupervised loss terms (Yamada et al. \citeyear{yamada2018paired}; Abramson et al. \citeyear{abramson2020imitating}; Özdemir, Kerzel, and
Wermter \citeyear{Oezdemir_2021_ICDL}), while large language models (LLMs) (Devlin et al. \citeyear{devlin2019bert}; Radford et al. \citeyear{radford2019language}; Brown et al. \citeyear{brown2020language}) introduced for various unimodal downstream language tasks rely on unsupervised learning for pretraining objectives.

In order to reduce this dependence on labeled data during training, we introduce a new training procedure, in which we limit the amount of training data used for supervised learning. More precisely, we only use a certain portion of training samples for crossmodal action-to-language and language-to-action translations whilst training unimodally on the rest of the training samples. As crossmodal translation requires each sample modality to be labeled with the other modality (e.g., an action sequence must be paired with a corresponding language description), we artificially simulate the realistic conditions where there is a large amount of unlabelled (unimodal) data but a much smaller amount of labeled (crossmodal) data.

\begin{wrapfigure}{R}{7cm}
    \includegraphics[width=7cm]{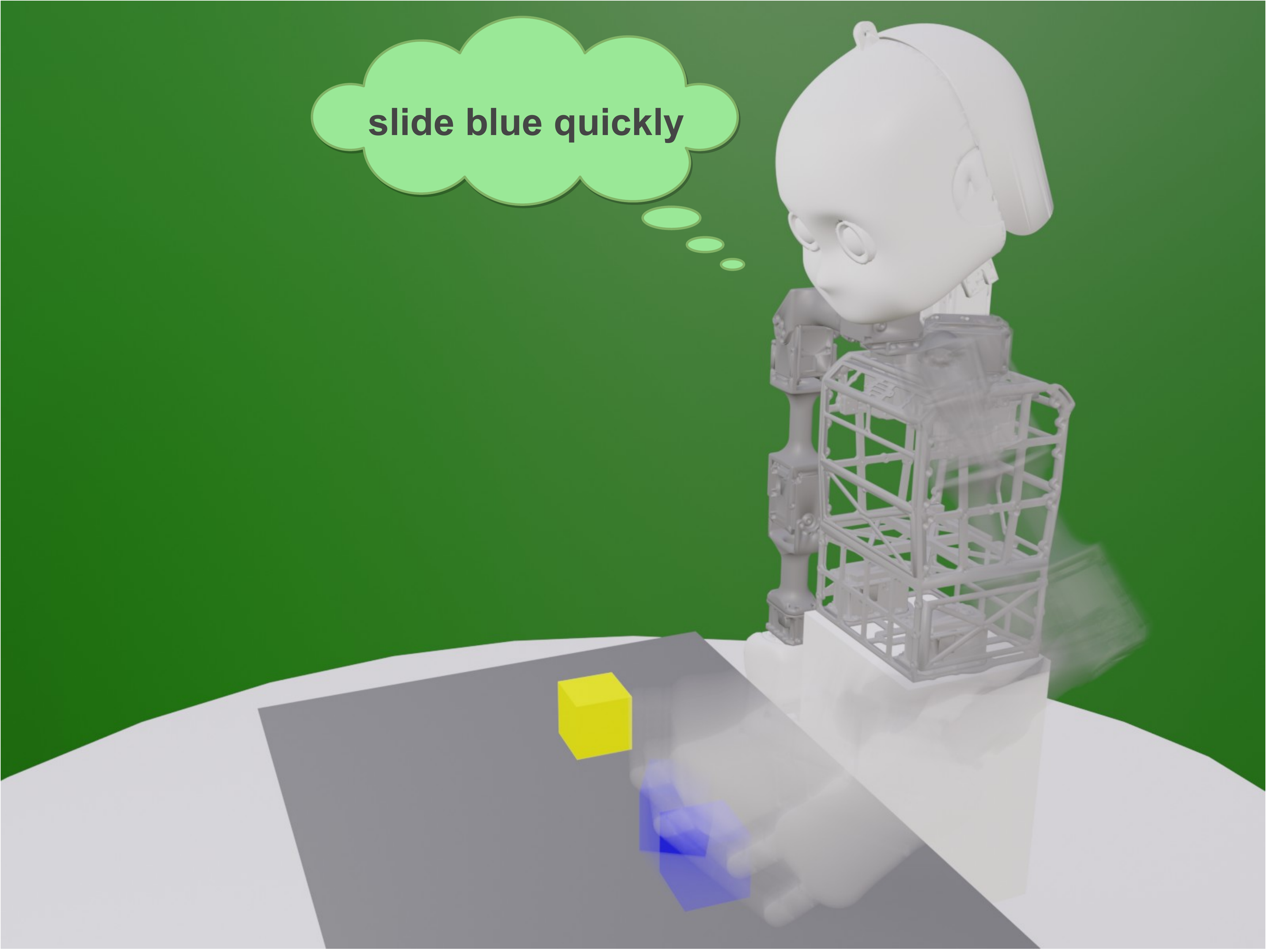}
    \caption{Our table-top object manipulation scenario in the simulation environment: the NICO robot is moving the blue cube on the table. The performed action is labeled as ``slide blue quickly''. Our approach can translate from language to action and vice versa; i.e., we perform actions that are described in language and also describe the given actions using language.}
    \label{fig:nico}
\end{wrapfigure}

Another aspect of human language learning is that it takes place in an environment and while using different modalities such as vision and proprioception. Concepts such as weight, softness, and size cannot be grounded without being in the environment and interacting with objects. Language learning approaches that use multiple modalities and take action in an environment into account are preferable to those that use a unimodal approach to process large amounts of text. A recent study \citep{canals_mor_2023} in language teaching concludes that learning is enhanced when the language learner uses language to produce meaningful outputs. Hence we strive to devise embodied multimodal models that tackle language grounding.
To this end, our robotic object manipulation dataset is generated from a simulation setup as seen in Figure \ref{fig:nico}. We use a humanoid child-size robot Neuro-Inspired COmpanion (NICO) (Kerzel et al. \citeyear{kerzel2017nico}; Kerzel et al. \citeyear{kerzel2020teaching}) to perform various actions on cubes on a table and label those actions with language descriptions. We introduce further details of our setup in Section 4.

Different from other approaches, our previous Paired Gated Autoencoders (PGAE) model \citep{pgae2022} can bidirectionally translate between language and action, which enables an agent not only to execute actions according to given instructions but also to recognize and verbalize its own actions or actions executed by another agent. As the desired translation task is communicated to the network through an additional signal word in the language input, PGAE can flexibly translate between and within modalities during inference. However, when trained under limited supervision conditions, PGAE performs poorly on the action-to-language translation task, under two conditions: Firstly, we experiment with reducing the number of supervised training {\em iterations} while using the whole data set for supervised training. Secondly, we experiment with reducing the number of training {\em samples} used with the supervised signals. In both instances, though the first is more trivial than the second, the action-to-language performance of PGAE suffers as the proportion of supervision decreases.

To overcome this hurdle, we present a novel model, Paired Transformed Autoencoders (PTAE), in this follow-up paper. Inspired by the successful application of the Crossmodal Transformer in vision-language navigation by the Hierarchical Cross-Modal Agent (HCM) architecture \citep{irshadhcm2021}, PTAE replaces PGAE's gated multimodal fusion mechanism and optionally the LSTM-based (long short-term memory) \citep{hochreiter1997long} encoders with a Crossmodal Transformer. Thanks to its more efficient and sequence-retaining crossmodal attention mechanism, PTAE achieves superior performance even when an overwhelming majority of training iterations (e.g., 98 or 99\%) consist of unsupervised learning. When the majority of training samples are used for unsupervised learning, PTAE still maintains its perfect action-to-language performance up to 80\% of training samples learned unimodally and performs relatively well for the 90\% case (over 80\% sentence accuracy). Even for the cases where only 1 or 2\% of the training samples are used in a supervised fashion, which is analogous to realistic few-shot learning settings, PTAE describes actions well over chance level with up to 50\% success rate.
Our results hint that PTAE precludes the need for large amounts of expensive labeled data, which is required for supervised learning, as the new architecture with the Crossmodal Transformer as the multimodality fusion technique significantly outperforms PGAE (Özdemir et al. \citeyear{pgae2022}) under the limited supervision training conditions.

Furthermore, inspired by the concept of incongruence in psychology and to test the robustness of the trained model to noise, for each task we introduce an extra input that is contradictory to the expected output of the model. For example, for language-to-action translation, we introduce extra conflicting action input showing an action that is different from the expected action from the model. The intertwined processing of language and action input in the Crossmodal Transformer resembles the tight interconnection between language and sensorimotor processes that has been observed in the human brain (Hauk, Johnsrude, and Pulvermüller \citeyear{hauk2004somatotopic}; van Elk et al. \citeyear{van2010functional}). Embodied accounts of human language comprehension assume that linguistic information induces mental simulations of relevant sensorimotor experiences. As a direct consequence of embodied language processing, conflicts between linguistic input and sensorimotor processes have been shown to result in bidirectional impairments of language comprehension on the one hand and perceptual judgments and motor responses on the other hand (Aravena et al. \citeyear{aravena2010applauding}; Glenberg and Kaschak \citeyear{glenberg2002grounding}; Kaschak et al. \citeyear{kaschak2005perception}; Meteyard, Bahrami, and Vigliocco \citeyear{meteyard2007motion}), although the strength of these behavioral effects has recently been debated (Winter et al. \citeyear{winter2022action}). In our PTAE model, we found asymmetry in terms of the impact of the action and language modalities on the performance of the model. Regardless of the output modality, introducing extra contradictory action input affects the model performance much more than introducing it in the language modality.

Our contributions in this work can be summarized as:
\begin{enumerate}
    \item We introduce PTAE that handles realistic learning conditions that mainly include unsupervised/unpaired language and action experiences while requiring minimal use of labeled data, which is expensive to collect.
    \item We show plausible behavior of the model when testing it with psychology-inspired contradictory information.
\end{enumerate}

The remainder of this paper is as follows: in Section 2, we summarize different approaches in language grounding with robotic object manipulation. In Section 3, we define our PTAE in detail. Section 4 introduces the experiments and their results. In Section 5, we discuss these results, while Section 6 concludes the paper.

\section{Related Work}
There are several approaches toward intelligent agents that combine language learning with interactions in a 3D environment.
% In 2020,
A comprehensive research program (Abramson et al. \citeyear{abramson2020imitating}) proposed %a complex paradigm
combining supervised learning, reinforcement learning (RL), and imitation learning. In the environment, two agents communicate with each other as one agent (setter) asks questions to or instructs the other (solver) that answers questions and interacts with objects accordingly. However, the scenario is abstract with unrealistic object interaction. Hence, proprioception is not used as the actions are high level, and a transfer of the approach from simulation to the real world would be non-trivial.

Jang et al. (\citeyear{bc-z}) proposed BC-Z which leverages a large multi-task dataset (100 tasks) to train a single policy, which is supervised with behavior cloning to match the actions demonstrated by humans in the dataset. To generalize to new tasks, the policy is conditioned on a task description; a joint embedding of a video demonstration, and a language instruction. This allows passing either the video command or the language command to the policy when being trained to match the actions in a demonstration. BC-Z generalizes to different tasks but requires a large collection of human demonstrations, which is expensive. It also relies on human intervention to avoid unsafe situations and to correct mistakes. 

Inspired by \cite{yamada2018paired}, we introduced the bidirectional Paired Variational Autoencoders (PVAE) \citep{Oezdemir_2021_ICDL} that is capable of modeling both language-to-action and action-to-language translation in a simple table-top setting where a humanoid robot interacts with small cubes. The approach can pair each robotic action sample (a sequence of joint values and visual features) with multiple language descriptions involving alternative words replacing original words. The two variational autoencoder networks of the model do not share any connections but are aligned with a binding loss term. Due to the lack of common multimodal representations, PVAE needs to be prepared for each translation task in advance. To overcome this issue, we proposed a bidirectional attention-based multimodal network, PGAE \citep{pgae2022}, which can flexibly translate between the two modalities with the help of a signal phrase.  

Another approach, CLIPort \citep{shridhar2021cliport},
combines the CLIP model (Radford et al. \citeyear{radford2021clip}) for pretrained vision-language representations with the Transporter model (Zeng et al. \citeyear{zeng2020transporter}) for robotic manipulation tasks. Transporter takes an action-centric approach to perception by detecting actions, rather than objects,
and then learns a policy, which allows CLIPort to exploit geometric symmetries for efficient representation learning. On multiple object manipulation tasks, CLIPort outperforms CLIP and Transporter alone. Further, CLIPort trained on multiple tasks performs better in most cases than CLIPort trained only on particular tasks. This supports the hypothesis that language-conditioned task-learning skills can be transferred from one task to another. However, the approach is only realized with a relatively simple gripper as it does not output joint angle values but 2D pixel affordance predictions. The actual action execution relies on the calibration between the robotic arm base and the RGB-D camera.

More recently, the same authors introduced Perceiver-Actor (PERACT) \citep{shridhar2022peract}, which is designed to efficiently learn multi-task robotic manipulations according to given language input by utilizing voxel grids extracted from RGB-D images. The backbone of the model is the Transformer-based Perceiver IO (Jaegle et al. \citeyear{perceiverio}) that uses latent vectors to tackle the processing of very long sequences. After the processing of appended language and voxel encodings by Perceiver IO, the voxels are decoded again to generate discrete actions by using linear transformations. PERACT achieves promising results in multiple tasks such as opening a drawer, turning a tap, and sliding blocks. However, as it only produces discrete actions, it relies on a random motion planner to execute instructions.

SayCan (Ahn et al. \citeyear{saycan2022arxiv}), utilizes LLMs to provide task-grounding capabilities to the agent, which is capable of executing short-horizon commands. The use of LLMs helps to ground these capabilities in the real world using value functions of the agent in order to produce feasible and useful instructions. However, the approach is limited to the set of skills that the agent can possess in the environment. An LLM is utilized to assign affordance probabilities to these skills according to a given high-level user instruction. The way these skills are defined in language (the wording, the length, etc.) can affect the performance of the whole system, e.g., LLMs tend to favor shorter phrases over longer ones.

GATO (Reed et al. \citeyear{gato2022}) is a single multi-task, multi-embodiment model that is general and performs well on hundreds of tasks in various domains such as playing Atari games, manipulating objects, image captioning, etc. Regardless of the modality (e.g., vision, proprioception, language, etc.), the input is flattened and embedded before it is provided to the model. The model is a large Transformer decoder that has the same weights and architecture for all tasks and is trained solely in a supervised manner. However, despite performing moderately in each task, the approach cannot compete with specialized approaches in various tasks.

The encoder-decoder-based VisuoMotor Attention model, VIMA for short, (Jiang et al. \citeyear{vima2022}) is another object manipulation approach. It deals with robot action generation from multimodal prompts by interleaving language and image or video frame tokens at the input level. VIMA uses an object detection module to extract objects and bounding boxes from visual input to use as object tokens. The object tokens are then interleaved with the language tokens and processed by the pretrained T5 model (Raffel et al. \citeyear{2020raffelt5}) which is used as the encoder. On the decoder end, the approach uses a causal Transformer decoder which consists of cross- and self-attention layers and autoregressively generates actions based on the history of previous actions and the multimodal prompt. It is shown that VIMA outperforms state-of-the-art approaches, including GATO, on a number of increasingly difficult object manipulation tasks involving zero-shot generalization with unseen objects and their combinations. An apparent weakness of VIMA is that it relies on the performance of off-the-self object detectors.

Different from most of the aforementioned approaches, our model is bidirectional: it can not only produce actions according to given language descriptions but also recognize actions and produce their descriptions. As our model is based on an autoencoder-like architecture, it can be trained in a mostly unsupervised way by asking the model to reproduce the given language or proprioception input. Moreover, our approach is flexible during inference since it does not need to be reconfigured for the translation task: due to the inclusion of the task signal in the language input, our PTAE can reliably execute the desired task on the go, whether it is a translation from language to action or vice versa. This is an essential step towards an autonomous agent that can interact within the environment as well as communicate with humans.

\section{Paired Transformed Autoencoder}
\begin{figure*}
  \includegraphics[width=\textwidth]{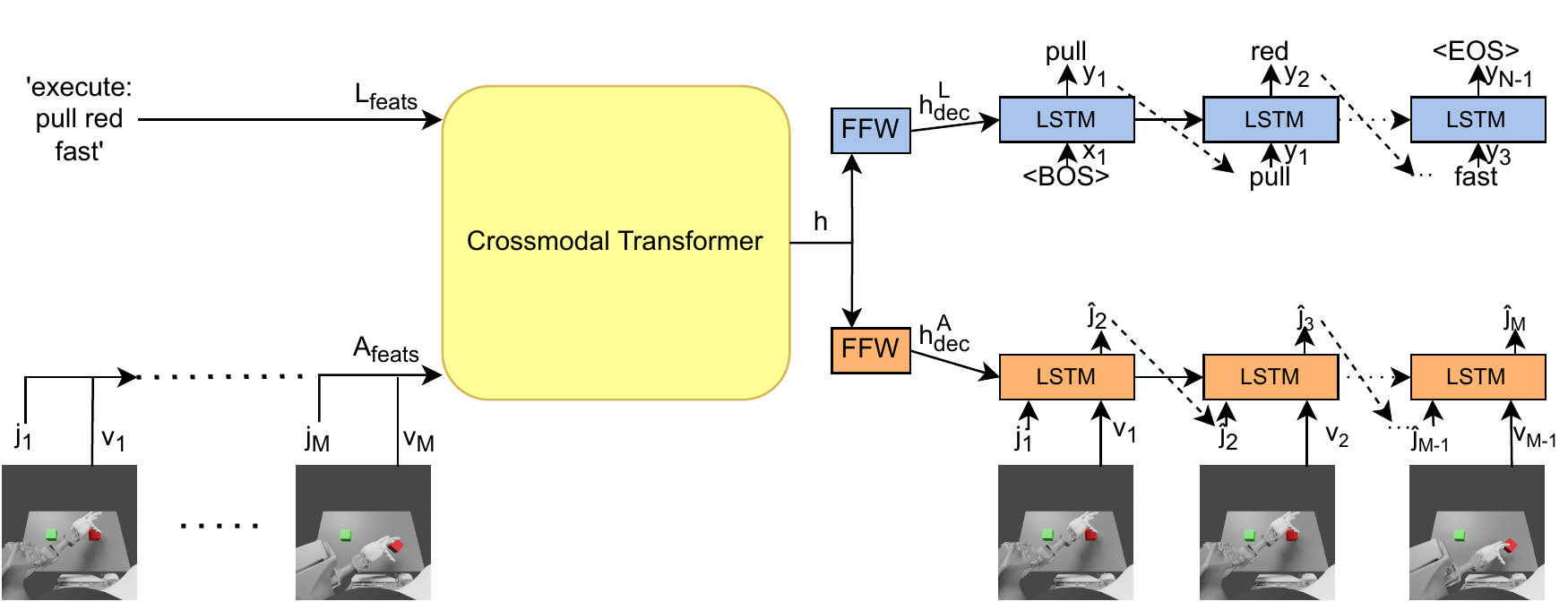}
  \caption{The architecture of the PTAE model. The inputs are a language description (incl. a task signal) and a sequence of visual features (extracted using the channel-separated convolutional autoencoder) and joint values, while the outputs are a description and a sequence of joint values. Language encoder can be an LSTM, the BERT Base model (Devlin et al. \citeyear{devlin2019bert}), or the descriptions can be directly passed to the transformer word by word. The action encoder can be an LSTM or the action sequence can be passed directly to the transformer. Both decoders are LSTMs - we show unfolded versions of the LSTMs. The bottleneck, where the two streams are connected, is based on the Crossmodal Transformer. h is the shared representation vector.} \label{modelarchtrans}
\end{figure*}
Our model, named PTAE, is an encoder-decoder architecture that is capable of bidirectional translation between robot actions and language. It consists of a Crossmodal Transformer that is the backbone and multimodality fusion mechanism of the architecture, and LSTM-based decoders that output language and joint values respectively. As input, PTAE accepts language descriptions of actions including the task signal, which defines the translation direction, as well as a sequence of the concatenation of multivariate joint values and visual features. According to the task signal, PTAE outputs joint values required for executing a particular action or it outputs language descriptions of an action.

As shown in Figure \ref{modelarchtrans}, PTAE is composed of a Crossmodal Transformer, which accepts multimodal input (i.e., language, proprioception, and vision), and language and action decoders that output language descriptions and joint values respectively. The language and action input can optionally be preprocessed by LSTM-based encoders as in the case of PGAE\footnote{For exact definitions of LSTM-based language and action encoder, readers may refer to the PGAE paper \citep{pgae2022}.}. However, after some initial trials with both cases, in this paper, we do not use any extra encoding layers before the Crossmodal Transformer for the sake of simplicity and model size as we do not see any significant change in the performance.

\subsection{Crossmodal Transformer}
\begin{figure*}
  \centerline{\includegraphics[width=0.55\textwidth]{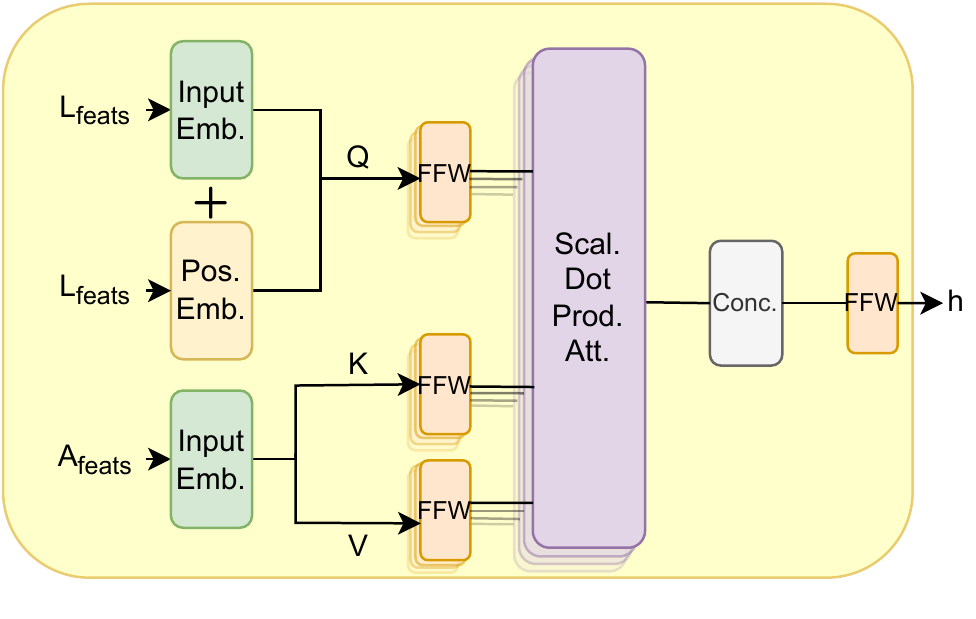}}
  \caption{The architecture of the Crossmodal Transformer: Language features are embedded and used as the query vector (Q), whereas the embedded action features are used as the key (K) and value (V) vectors. The positional embedding is applied only to the language features. The multi-head attention (MHA) involves the Q-, K- and V-specific feedforward (FFW) and scaled dot product attention layer following the original Transformer architecture. The multiple heads are then concatenated and fed to the final FFW, which outputs the common hidden representation vector h.} \label{crosstransformer}
\end{figure*}

The Crossmodal Transformer replaces the Gated Multimodal Unit (GMU) (Arevalo et al. \citeyear{arevalo2020gated}) in our previous PGAE model \citep{pgae2022} and can be employed essentially as language and action encoders. The simplified architecture of the Crossmodal Transformer can be seen in Figure \ref{crosstransformer}. The functionality of the Crossmodal Transformer is to extract the common latent representations of paired language and action sequences. Following the HCM architecture \citep{irshadhcm2021}, we use the language modality as queries ($Q$ vectors) and the action modality (concatenated visual features and joint values) as keys ($K$ vectors) and values ($V$ vectors). The language descriptions are represented as one-hot encoded vectors, whilst action input is composed of joint values of NICO's left arm and the visual features from images recorded by the camera in NICO's eye. As in PGAE, we use a channel-separated convolutional autoencoder (CAE) to extract visual features from images. The Crossmodal Transformer encodes the common latent representations as follows:
\begin{align*}
Q = \text{ReLU}\left ( W^{token}\cdot x_{t} + b^{token} \right ) + \text{PE}(x_{t}) \hspace{5mm} (1\leq t\leq N+1),\\
K, V = \text{ReLU}\left ( W^{act}\cdot \left [v_{t};j_{t} \right ] + b^{act} \right ) \hspace{5mm} (1\leq t\leq M), \\
A_{t} = \text{MHA}(Q, K, V) \hspace{5mm} (1\leq t\leq N+1),\\
h_{t} = \text{PWFF}(A_{t}) \hspace{5mm} (1\leq t\leq N+1), \\
h = \text{AvgPool}(h_{t}) \hspace{5mm} (1\leq t\leq N+1),  
\end{align*}
where $x$, $v$, and $j$ are linguistic, visual, and proprioceptive inputs respectively -- note that when no language or action encoder is used, $x$ corresponds to $\textrm{L}_{\textrm{feats}}$, while the concatenation of visual features and joint values $\left [v_{t};j_{t} \right]$ corresponds to $\textrm{A}_{\textrm{feats}}$ in Figure \ref{crosstransformer}. ReLU is the rectified linear unit activation function while PE, MHA, and PWFF are the positional encodings, multi-head attention layer, and the position-wise feedforward layer as used in the original Transformer paper (Vaswani et al. \citeyear{vaswani2017attention}). As the Transformer architecture does not include any recurrence, we employ a fixed sinusoid function-based PE layer on the language features to include the position information. $A_{t}$ is the crossmodal attention vector for time step $t$, whereas $h_{t}$ is the hidden vector for time step $t$.  AvgPool is the average pooling applied on the time axis to the sequential hidden vector to arrive at the common latent representation vector $h$. For our experiments, we employ a single-layer Crossmodal Transformer with 4 parallel attention heads.

\subsection{Language Decoder}
We use an LSTM as the language decoder in order to autoregressively generate the descriptions word by word by expanding the common latent representation vector $h$ produced by the Crossmodal Transformer:
\begin{align*}
    h_{0}^{\text{dec}}, c_{0}^{\text{dec}} &= W^{\text{dec}} \cdot  h + b^{\text{dec}}, \\
    h_{t}^{\text{dec}}, c_{t}^{\text{dec}} &= \text{LSTM}(y_{t-1}, h_{t-1}^{\text{dec}}, c_{t-1}^{\text{dec}}) \hspace{3mm} (1\leq t\leq N-1), \\
    y_{t} &= \text{soft}( W^{\text{out}} \cdot h_{t}^{\text{dec}} + b^{\text{out}}) \hspace{5mm} (1\leq t\leq N-1),
\end{align*}
where $\text{soft}$ represents the softmax activation function. $y_{0}$ is the vector for the symbol indicating the beginning of the sentence, the $<$BOS$>$ tag. 

\subsection{Action Decoder}
Similarly, an LSTM is employed as the action decoder to output joint angle values at each time step with the help of the common representation vector $h$:
\begin{align*}
    h_{0}^{\text{dec}}, c_{0}^{\text{dec}} &= W^{\text{dec}} \cdot  h + b^{\text{dec}},\\
    h_{t}^{\text{dec}}, c_{t}^{\text{dec}} &= \text{LSTM}(v_{t}, \hat{\jmath}_{t}, h_{t-1}^{\text{dec}}, c_{t-1}^{\text{dec}})\hspace{5mm}(1\leq t\leq M-1), \\
    \hat{\jmath}_{t+1} &= \text{tanh}( W^{\text{out}} \cdot h_{t}^{\text{dec}} + b^{\text{out}}) \hspace{5mm} (1\leq t\leq M-1),
\end{align*}
where $\hat{\jmath}_{t}$ is the predicted joint values for time step t and $\text{tanh}$ is the hyperbolic tangent activation function. We take $\hat{\jmath}_{1}$ as ${j}_{1}$, i.e.,\ ground-truth joint angle values corresponding to the initial position of the arm. The visual features used as input $v$ are extracted from the ground-truth images and used similarly to teacher forcing, whereas the joint angle values $\hat{\jmath}_{t}$ are used autoregressively.

\subsection{Visual Feature Extraction}
Following the PGAE pipeline \citep{pgae2022}, the channel-separated convolutional autoencoder (CAE) is used to extract visual features from first-person images from the eye cameras of NICO recorded in the simulation. We utilize channel separation when extracting visual features: an instance of the CAE is trained for each RGB color channel. In a previous paper \citep{Oezdemir_2021_ICDL}, we show that channel separation distinguishes object colors more accurately than the regular CAE without channel separation. 

We feed each instance of the channel-separated CAE with the corresponding channel of RGB images of size $120 \times 160$. The channel-separated CAE is made up of a convolutional encoder, a fully-connected bottleneck, and a deconvolutional decoder. Each RGB channel is trained separately, after which we extract the channel-specific visual features from the bottleneck and concatenate them to arrive at composite visual features. These visual features make up $v$ which is used as vision input to PTAE. For further details on the visual feature extraction process, readers may refer to \cite{Oezdemir_2021_ICDL}.

\subsection{Loss Function}
We use two loss functions to calculate the deviation from the ground-truth language descriptions and joint values. The language loss, $L_{\text{lang}}$, is calculated as the cross entropy between input and output words, while the action loss, $L_{\text{act}}$, is the mean squared error (MSE) between original and predicted joint values:
\begin{equation*}
L_{\text{lang}} = \frac{1}{N-1} \sum_{t=1}^{N-1}\left ( -\sum_{i=0}^{V-1} x_{t+1}^{\left [ i \right ]}\log y_{t}^{\left [ i \right ]}\right),
\end{equation*}
\begin{equation*}
L_{\text{act}} = \frac{1}{M-1} \sum_{t=1}^{M-1}\left \| j_{t+1} - \hat{\jmath}_{t+1}  \right \|_{2}^{2},
\end{equation*}
where $V$ is the vocabulary size, $N$ is the number of words per description, and M is the sequence length for action trajectories. The total loss is then the sum of the language and action losses:
\begin{equation*}
L_{\text{all}} = \alpha L_{\text{lang}} + \beta L_{\text{act}}
\end{equation*}
where $\alpha$ and $\beta$ are weighting factors for language and action terms in the loss function. In our experiments, we take both $\alpha$ and $\beta$ as 1.0. We use the identical loss functions as PGAE except for the weight vector used in the language loss to counter the imbalance in the frequency of words, after seeing that it is unnecessary for PTAE.
\subsection{Training Details}
Visual features are extracted in advance by the channel-separated CAE before training PTAE and PGAE. Visual features are necessary to execute actions according to language instructions since cube arrangements are decisive in manipulating the left or right object, i.e., determining whether to manipulate the left or right cube depends on the position of the target cube. After extracting visual features, both PGAE and PTAE are trained end-to-end with all three modalities. After initial experiments, PGAE is trained for 6,000 epochs, while PTAE is trained for 2,500 epochs using the gradient descent algorithm and Adam optimizer \citep{kingma2015adam}. For PTAE, we decided that $h$ has 256 dimensions following \cite{irshadhcm2021}, whereas the same vector has 50 dimensions in PGAE. $x$ has 28 dimensions, $j$ has 5 dimensions, $N$ is equal to 5, while $M$ is 50 for fast and 100 for slow actions. For both PGAE and PTAE, we take the learning rate as $10^{-5}$ with a batch size of 6 samples after determining them as optimal hyperparameters. PTAE has approximately 1.5M parameters compared to PGAE's a little over 657K parameters.

\section{Experiments}
We use the same dataset \citep{Oezdemir_2021_ICDL} as in the PGAE paper \citep{pgae2022}, except 
that in this paper we exclude experiments with another agent from the opposite side of the table. The dataset encompasses 864 samples of sequences of images and joint values alongside their textual descriptions. It consists of robot actions on two cubes of different colors on the table by the NICO robot, generated using inverse kinematics and created in the simulation environment using Blender software\footnote{\url{https://www.blender.org/}}. The NICO robot has a camera in each eye, which is used to record a sequence of egocentric images. According to the scenario, NICO manipulates one of the two cubes on the table with its left arm at a time. Accordingly, we use and record 5 joints of the left arm during object manipulation. In total, the dataset includes 12 distinct actions\footnote{The actions are distinguished based on the action type (PUSH, PULL, or SLIDE), the position of the manipulated object (LEFT, or RIGHT), and speed (SLOW, or FAST).}, 6 cube colors, 288 descriptions \footnote{As we have 6 action words, 12 color words, and 4 speed words, we reach 288 distinct descriptions.}, and 144 patterns\footnote{We have 12 distinct actions and 12 cube arrangements (e.g., red-green); thus their combinations make 144.} (action \& cube arrangement combinations). The 144 patterns are randomly varied six times in terms of action execution in simulation: we arrive at a dataset of 864 samples in total. Out of 864 samples, 216 samples that involve every unique description and action type are excluded and used as the test set. The remaining 648 samples make up the training set. The vocabulary consists of the following words divided into 3 categories:
\begin{itemize}
    \item 6 action words (3 original/3 alternative): ``push/move-up'', ``pull/move-down'', ``slide/move-sideways''
    \item 12 color words (6 original/6 alternative): ``red/scarlet'', ``green/harlequin'', ``blue/azure'', ``yellow/blonde'', ``cyan/greenish-blue'', ``violet/purple'' 
    \item 4 speed words (2 original/2 alternative): ``slowly/unhurriedly'', ``fast/quickly''
\end{itemize}
The sentences consist of a word from each category: therefore, our textual descriptions are 3-word sentences. For more details on the dataset, readers may consult our previous work \citep{Oezdemir_2021_ICDL}. PGAE and PTAE are trained on this dataset and their performances are tested in terms of action-to-language and language-to-action translations under different amounts of supervision.

\paragraph*{Task signals.}

We use four signals to train PTAE. According to the given signal, the input and output of the model change. The signals are:
\begin{itemize}
  \item \textbf{Describe}: action-to-language translation
  \item \textbf{Execute}: language-to-action translation
  \item \textbf{Repeat Action}: action-to-action translation
  \item \textbf{Repeat Language}: language-to-language translation
\end{itemize}
According to the latter two ``repeat'' signals, the network uses mainly unimodal information. The ``describe'' and ``execute'' signals, on the other hand, involve crossmodal translation from one modality to the other. The unimodal signals are used in the unsupervised learning of an autoencoder, whereas the crossmodal signals are used in supervised learning, where coordinated action values and language labels must be available.
In the case of PGAE training, an additional ``repeat both'' signal is also used, which also requires coordinated labels, and leads to slightly better performance \citep{pgae2022}. For the PTAE, however, this was found unnecessary.

\paragraph*{Reduction of supervised training.}

\begin{figure*}[h]
    \centering
    \includegraphics[width=1\textwidth]{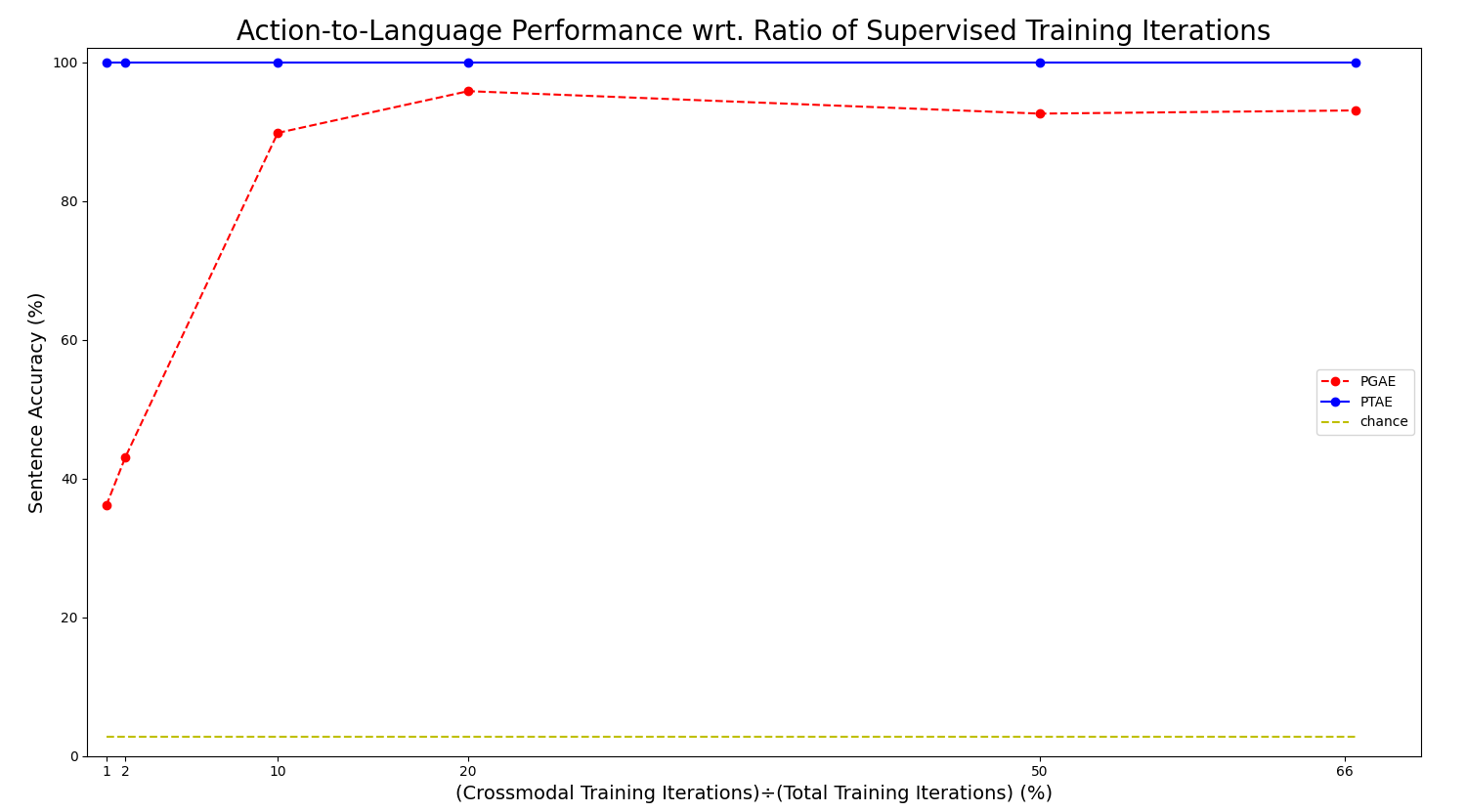}
    \caption[Action-to-language translation performance on the test set wrt. supervised training iterations]{Sentence accuracy for action-to-language translation on the test set wrt. supervised training iterations. Supervised training refers to crossmodal translation cases ``describe'' and ``execute''. The two crossmodal signals receive the same number of iterations between them out of the supervised iterations. We report the results for 1\%, 2\%, 10\%, 20\%, 50\%, and 66.6\% (the regular training case) crossmodal (supervised) iterations. These percentages correspond to the fraction of supervised training iterations for PGAE and PTAE. Note that the 100\% case is not shown here, since the models need unsupervised iterations (unimodal repeat signals) to be able to perform the ``repeat language'' and ``repeat action'' tasks.
    }
    \label{fig:sentence_acc_iter}
\end{figure*}

 We restrict the amount of supervision by increasing the ratio of unsupervised learning iterations, i.e., training with the unimodal ``repeat'' signals, in the overall training iterations. Thereby the ratio of supervised learning iterations, i.e., training with the crossmodal signals, decreases. The resulting training paradigm is analogous to developmental language learning, where an infant is exposed only to a limited amount of supervision. We train both PTAE and PGAE with varying ratios of unimodal/total training iterations. For another set of experiments, we restrict the amount of supervision by limiting the proportion of training samples used for crossmodal translation tasks. We test the performance of both models with varying degrees of unsupervised training under different schemes (limiting the percentage of \emph{iterations} or \emph{samples}) on the crossmodal translation tasks. 

\begin{figure*}[h]
    \centering
    \includegraphics[width=1\textwidth]{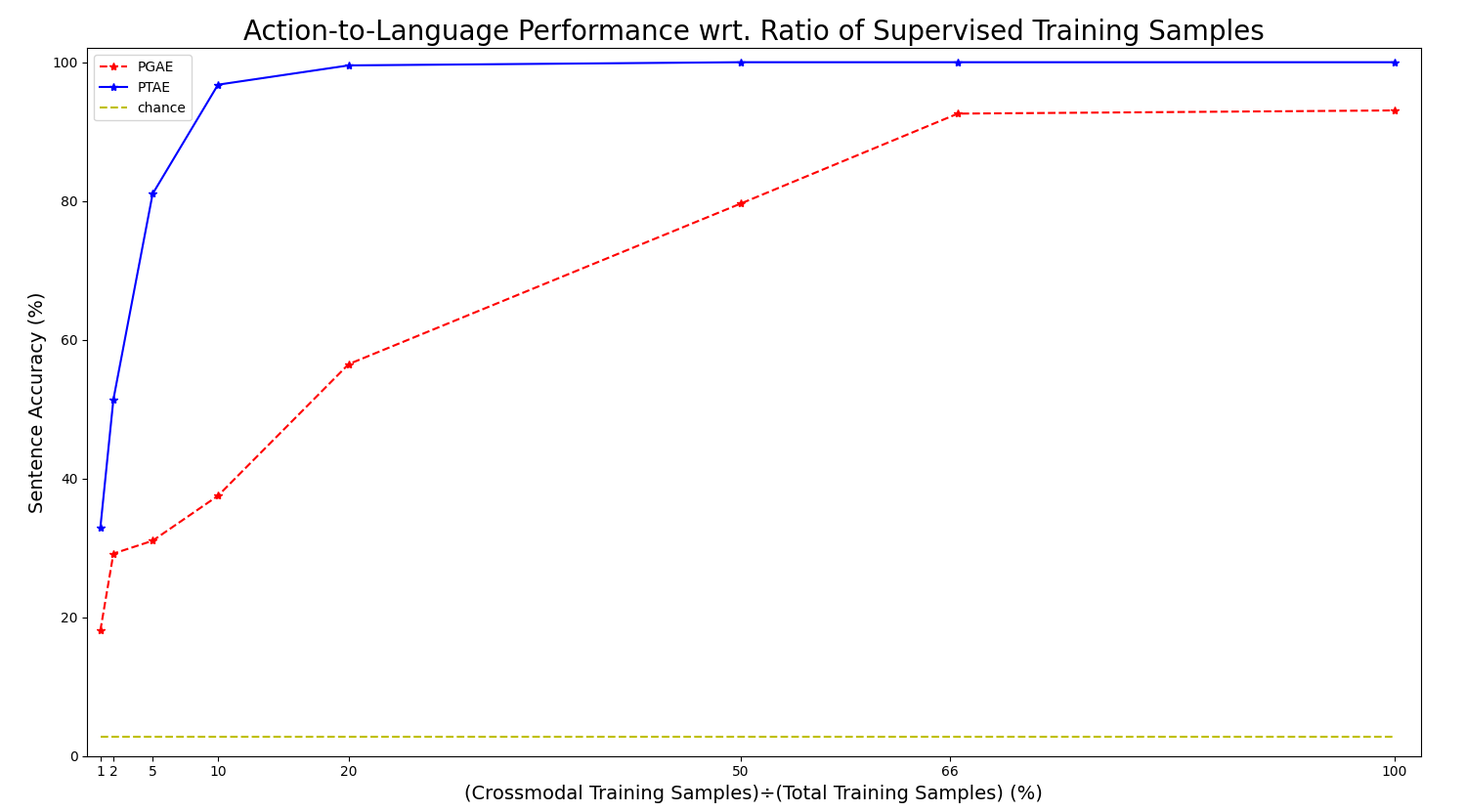}
    \caption[Action-to-language translation performance on the test set wrt. supervised training samples]{Sentence accuracy for action-to-language translation on the test set wrt. supervised training samples. Supervised training refers to crossmodal translation cases ``describe'' and ``execute''. We limit the number of training samples for the supervised tasks. We report the results for the 1\%, 2\%, 5\% 10\%, 20\%, 50\%, and 66.6\% cases as well as the 100\% regular training case. These percentages correspond to the fraction of training samples used exclusively for the supervised training for PGAE and PTAE, i.e., both ``execute'' and ``describe'' signals are trained with only a limited number of samples corresponding to the percentages.}
    \label{fig:sentence_acc_data}
\end{figure*}

In this work, we investigate action-to-language and language-to-action translations because they are the more important and difficult tasks. For the ``repeat'' tasks, the results match our previous work; therefore, the readers can refer to our publication \citep{pgae2022}. Figure \ref{fig:sentence_acc_iter} shows the results of PGAE and PTAE on action-to-language translation with different percentages of training {\em iterations} used in a supervised fashion. Both PGAE and PTAE with different training regimes based on different proportions of supervised training iterations achieve accuracies higher than the chance level (2.78\%), which we calculate based on our grammar (action, color, speed): $1\div(3\times6\times2)$. The action-to-language translation performance of PGAE falls when the ratio of crossmodal (viz. supervised) training iterations is low, particularly when 10\% or a smaller proportion of the iterations are supervised. Even though the description accuracy slightly increases to over 95\% when supervised training amounts to only 20\% of all training iterations (it may partially be due to overfitting), it sharply drops to well below 50\% when the rate is decreased to 2\%. PGAE is able to describe 36\% of the test samples when only 1\% of the training iterations are used to learn crossmodal translations between action and language. In contrast, PTAE maintains its perfect description accuracy even when it has only been trained with 1\% supervised training iterations. While there is a detrimental impact of reduced supervision, i.e., the limitation on the percentage of crossmodal training iterations, on the action-to-language translation performance of PGAE, transformer-based PTAE is not affected by the same phenomenon. For space reasons, we do not report language-to-action results wrt. different percentages of supervised iterations, but we observed a similar trend comparable with Figure \ref{fig:sentence_acc_iter}.

In order to further investigate the performance of PTAE with limited supervision, we introduce a more challenging training regime. We limit the number of training {\em samples} shown to supervised signals, ``describe'' and ``execute'', and show the rest of the training samples only on ``repeat action'' and ``repeat language'' modes. We train both PGAE and PTAE with varying percentages of supervised training samples. The results can be seen in Figure \ref{fig:sentence_acc_data}. In all cases with different proportions of supervised training samples, both PGAE and PTAE outperform the chance level. While maintaining perfect sentence accuracy down to 20\% supervised training and keeping up its performance for 10\% supervised training for the ``describe'' signal, PTAE's performance drops sharply when the ratio of training samples used for crossmodal signals is 2\% and below. Nevertheless, PTAE beats PGAE in each case when trained on different percentages of supervised training samples. PGAE's performance suffers even when 50\% of training samples are used for supervised signals; it drops below 80\% - PTAE retains 100\% for the same case. It takes more than 90\% of the training samples to be exclusively used in the unsupervised signals for PTAE's performance to decrease meaningfully (from 100\% to 81\%), while this ratio is much lower for PGAE as its performance already drops significantly at 50\%. Even for 1\% supervised training samples which amount to only 7 training samples, PTAE manages to translate one-third of the test samples from action to sentences. 

\begin{figure*}[h]
    \centering
    \includegraphics[width=1\textwidth]{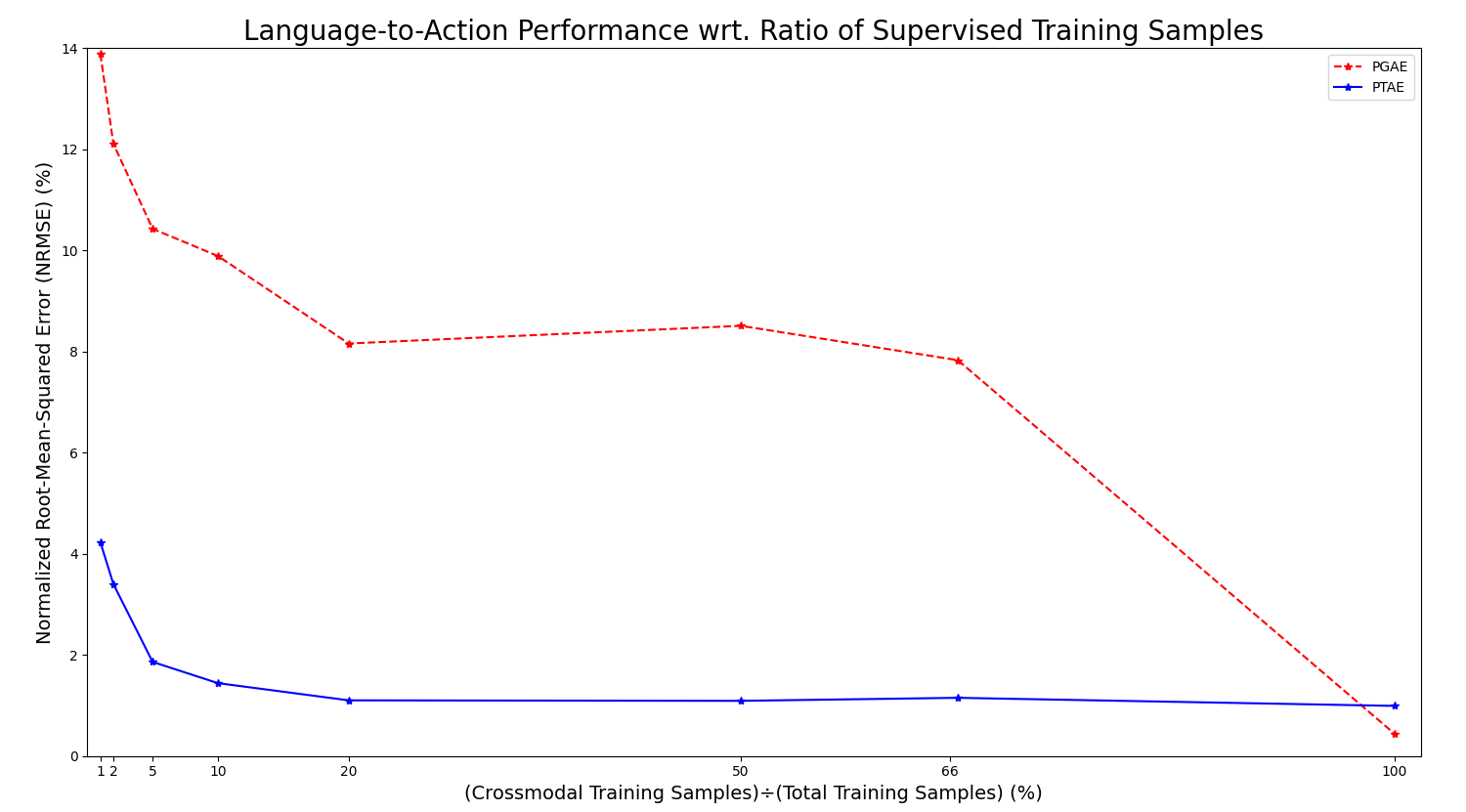}
    \caption[Language-to-action translation performance on the test set wrt. supervised training samples]{Joint value prediction error in language-to-action translation on the test set wrt. supervised training samples. Supervised training refers to crossmodal translation cases ``describe'' and ``execute''. We limit the number of training samples for the supervised tasks. We report the results for the 1\%, 2\%, 5\% 10\%, 20\%, 50\%, and 66.6\% cases as well as the 100\% regular training case. These percentages correspond to the fraction of training samples used exclusively for the supervised training for PGAE and PTAE. ``execute'' and ``describe'' translations are shown the same limited number of samples.}
    \label{fig:joint_error_data}
\end{figure*}

Language-to-action translation results with respect to different percentages of supervised training samples for PGAE and PTAE are shown in Figure \ref{fig:joint_error_data}. We show the deviation of the produced joint values from the original ones in terms of the normalized root-mean-squared error (NRMSE), which we obtain by normalizing the root-mean-squared error (RMSE) between the predicted and ground-truth values by the range of joint values -- the lower percentages indicate better prediction (0\% NRMSE meaning predicted values are identical with ground-truth values), whereas the higher percentages indicate worse prediction (100\% NRMSE meaning the RMSE between predicted and ground-truth values is equal to the range of possible values). We can see a similar trend as in action-to-language translation apart from the regular case (100\%) when PGAE has a lower error than PTAE, which is probably due to the fact that PGAE is trained for more than two times the number of iterations than PTAE since it takes longer for PGAE's training loss to reach a global minimum. In all other cases, limiting the ratio of training samples to be used in the supervised modes impacts PGAE's language-to-action performance heavily: the NRMSE rises from less than 0.5\% to almost 8\% when the percentage of supervised samples is reduced to two-thirds of the training samples. The error rate increases further as the number of training samples used in the crossmodal training modes decreases. The NRMSE for PTAE is also inversely proportional to the ratio of supervised training samples. However, the impact of limiting the number of training samples for supervised modes on PTAE is much lower than on PGAE. When the percentage of supervised training samples is reduced to 1\%, the deviation from the ground-truth joint values is only a little more than 4\% for PTAE, whereas the same statistic for PGAE is almost 14\%.   

\begin{figure*}[h]
    \centering
    \includegraphics[width=1\textwidth]{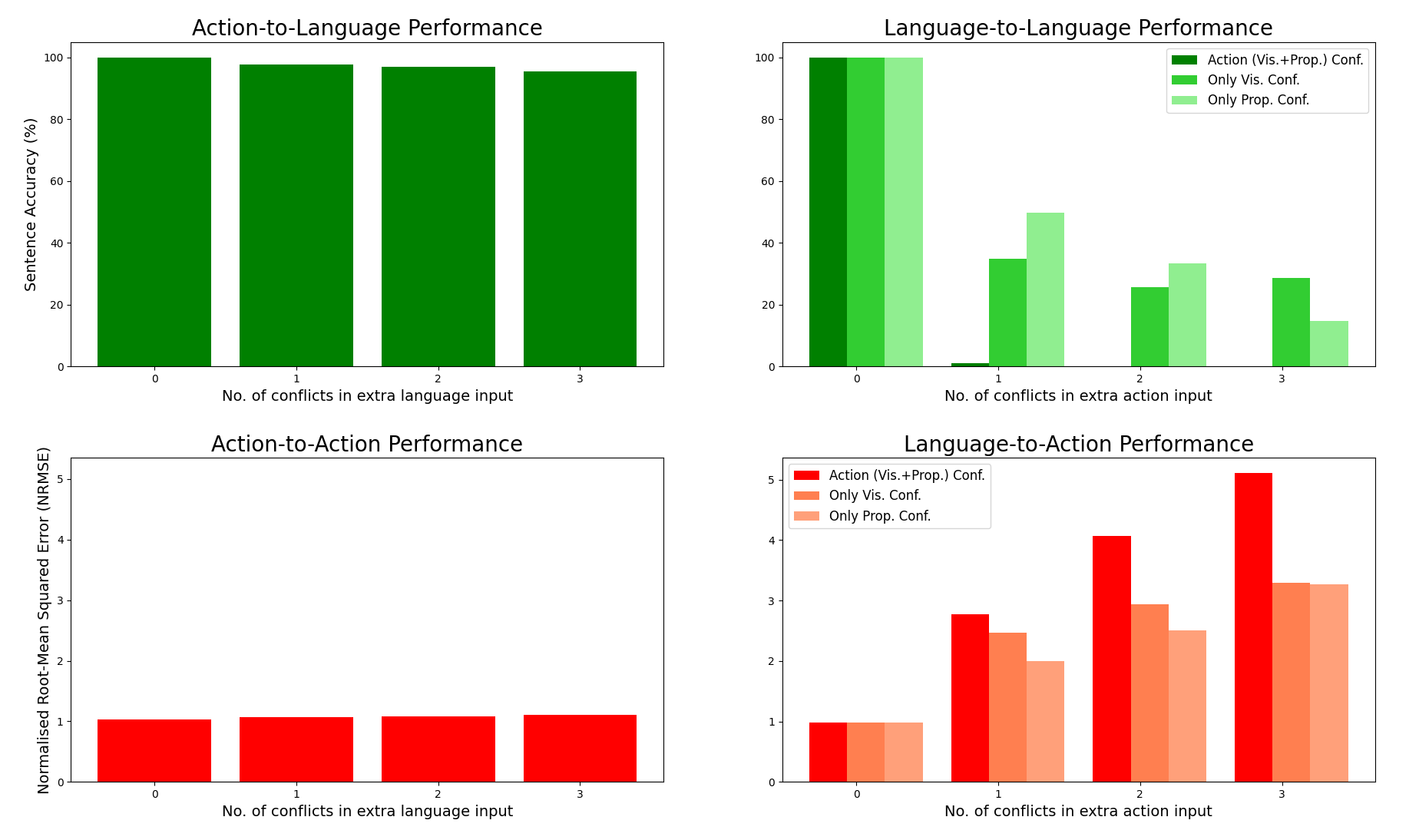}
    \caption[Model performance on the test set wrt. no. of conflict]{Model performance on the test set wrt.\ no.\ of conflicts introduced in the extra input. For action-to-language and language-to-language (the top row), we show the predicted sentence accuracies. For language-to-action and action-to-action, we show the normalized root-mean-squared error (NRMSE) for predicted joint values. The modality in which the conflicts are introduced is given in the x-axis. For each signal, we add extra conflicting inputs either in the action or language input. When the conflict is introduced in action, we also test having the conflict only in the vision and only in the proprioception submodality - in this case, the other submodality has the matching input.
    }
    \label{fig:performance_with_conf}
\end{figure*}

\paragraph*{Exposure to conflicting input modalities.}

We also investigate the impact of contradictory extra input on the performance of PTAE. For this, we use PTAE-regular that is trained with 33\% unsupervised training iterations and no contradictory input. We test the robustness of our approach to varying numbers of conflicts (up to 3) in the extra input. The definitions of the added conflict per task signal are: 
\begin{itemize}
    \item ``describe'': Here, we add a conflicting description to the language input (conflict in language).
    \item ``execute'': Here, we use a conflicting sequence of vision and proprioception input (conflict in action). 
    \item ``repeat action'': Here, we add a conflicting description to the language input (conflict in language).
    \item ``repeat language'': Here, we use a conflicting sequence of vision and proprioception input (conflict in action). 
\end{itemize}
The conflicts are introduced using the following scheme:
\begin{itemize}
    \item for the conflict in the extra language input; one, two, or all of the action, color, and speed words that constitute a description, do not match with those of the ground-truth paired description of the action. For instance, for the input action paired with the description ``push red slowly'', a description like ``push green slowly'' (one conflict present, namely color), or ``pull green fast'' (all three conflicts present; action, color, speed) is given to the model as conflicting extra language input.
    \item for the conflict in the extra action input; one, two, or all of the action-type, position, and speed aspects, which form distinct actions, do not match with the language description. We choose one of those action trajectories that are not paired with the given language input. The conflict(s) can be in the action type (e.g., pushing instead of pulling), the position of the manipulated object (e.g., the left cube being pulled instead of the right), or the speed of the action (e.g., the cube is being pulled fast instead of slowly).
\end{itemize}

The results of this experiment are given in Figure \ref{fig:performance_with_conf}. In the case of the ``describe'' and ``repeat action'' signals, the action supplies the relevant input whereas the language is the conflicting distractor. Here, we observe only a slight decrease in performance. In the case of action-to-language translation (``describe'') the sentence accuracy goes down from 100\% to 95\% when there are three conflicting input elements (action type, color, speed). Action-to-action (``repeat action'') translation manages to retain its performance as the error in joint values only slightly increases from 1.03\% to  1.09\% for the case with 3 conflicts.

In the case of ``execute'' and ``repeat language'' signals, the language supplies the relevant input while the action is the conflicting distractor. Here, we observe a big performance drop. Language-to-action translation (``execute'') suffers heavily as the deviation of the predicted joint values from the ground-truth joint values increases from 0.99\% to 4.95\%. In the language-to-language translation case (``repeat language''), PTAE loses its ability to repeat the given language description when one or more conflicting elements (action type, position, speed) are introduced with the extra input: the sentence accuracy decreases from 100\% to 0\%.

Therefore, we can see the asymmetric impact of conflicts in the two modalities, namely, when language input is introduced as a contradictory element, the performance drops slightly, whereas when the contradictory input is introduced in the action stream, the model is affected heavily and performs poorly. The output modality has no significant impact on the result; for example, we can see that both ``describe'' and ``repeat language'' output language at large, but they are affected very differently by the conflicting input. To test whether the bigger impact of conflicting action input is due to the involvement of two modalities in action (vision and proprioception), we also tried introducing the conflict either only in vision or only in proprioception (the relatively brighter bars in the two charts on the right in Figure \ref{fig:performance_with_conf}). In either case, the performance is still substantially negatively affected, although the drop in performance is naturally not as severe as introducing the conflict in both modalities.

\section{Discussion}

The experimental results on action-to-language and language-to-action translations show the superior performance and efficiency of our novel PTAE model under limited supervision. Limiting the percentage of supervised crossmodal iterations during training has no adverse effect on PTAE as it maintains its perfect sentence accuracy when translating from action to language. In contrast, the previous PGAE model's action-to-language translation accuracy drops substantially when only a tiny proportion of the training iterations are supervised. When we challenge both models more by limiting the number of training samples for the supervised crossmodal ``execute'' and ``describe'' signals, we see a similar pattern: when half or less than half of the training samples are used for supervised signals, action-to-language sentence accuracy for PGAE decreases directly proportional to the ratio of supervised samples. PTAE, on the other hand, retains its action-to-language performance up until when an overwhelming majority of training samples are used in a supervised fashion. Even after being trained with 2\% supervised training, which amounts to only 13 samples out of 648, PTAE is able to describe more than half of the action sequences correctly. All in all, PTAE shows superior action-to-language performance than PGAE for varied levels of limited supervision.

The adverse effect of limiting the number of supervised training samples on the language-to-action performance can already be seen for PGAE even when only one-third of the samples are excluded as the error rate between predicted and ground-truth joint values rises significantly. It continues to increase gradually after reducing the level of supervision further. On the contrary, PTAE is robust against limited supervision with respect to the ratio of crossmodal training samples until the supervised percentage is brought down heavily. Achieving similar error rates on the range from one-fifth of training samples to all of them being trained in a supervised fashion also shows that for PTAE the learning of language-to-action translation reaches a plateau, where added labels do not provide additional useful information. After reducing the supervised ratio further, it can be seen that the error rate gradually increases, albeit only just over 4\% for PTAE when only 7 samples are used for the supervised signals. Overall, these results indicate the clear superiority of Transformer-based multimodal fusion over a simpler attention mechanism by GMU in terms of performance and efficiency. Although it is relatively larger than PGAE, PTAE is trained much faster and reaches a global optimum in less than half of the training iterations of PGAE. It is clear from these results that scaled dot-product attention, which forms the backbone of the Crossmodal Transformer, can work with a low proportion of supervision during training, whereas gated attention, which is used by GMU, requires a much larger supervised proportion to learn the crossmodal mapping between action and language. The Crossmodal Transformer utilizes a relatively long set of matrix operations over all time steps (temporal information is kept until the extraction of the representation vector), while GMU relies on simpler equations over the mean input features that no longer bear a temporal dimension.

When introducing a conflicting modality input during testing, we observed an asymmetry in that a conflicting action input leads to a larger disturbance than a conflicting language input. One possible reason is that the Crossmodal Transformer architecture is asymmetric: As input, we are using action input as two input vectors (K and V: keys and values), whereas language as one input vector (Q: queries). This setting was chosen because the opposite setup (with action as queries) was found less performant. Our setup can be interpreted as language-conditioned action attention. A computationally more expensive architecture could combine both asymmetric setups, as has been done for learning vision and language representations (Lu et al. \citeyear{lu2019vilbert}).

Another possible reason for the larger impact of a conflicting action could be that the action input combines two submodalities, vision, and proprioception, and therefore involves more information than the language input. However, limiting the conflict to one of the submodalities did not completely remove the asymmetry as introducing the conflict only in one action submodality (vision or proprioception) still had a stronger effect on the model performance than a conflicting language input. Unlike language, vision contains the complete information to perform a task. Consider the example ``pull red slowly" for language-to-action translation. Here, the language does not contain any information about whether the object is on the left or right side, so the agent can only execute this correctly when also taking visual input into account during action execution. In contrast, in the opposite direction (action-to-language translation) and in action repetition, the visual input has complete information.

\section{Conclusion}
In this paper, we introduced a paired Transformer-based autoencoder, PTAE, which we trained largely by unsupervised learning with additional, but reduced supervision. The PTAE achieves significantly better action-to-language and language-to-action translation performance under limited supervision conditions compared to the former GMU-based model, PGAE. Furthermore, we tested the robustness of our new approach against contradictory extra input. In line with the concept of incongruence in psychology, these experiments show that conflict deteriorates the output of our model, and more conflicting features lead to higher interference. We also found an asymmetry between the action and language modalities in terms of their conflicting impact: the action modality has significantly more influence over the performance of the model regardless of the main output modality.

Our novel bidirectional embodied language learning model is flexible in performing multiple tasks and it is efficient and robust against the scarcity of labeled data. Hence, it is a step towards an autonomous agent that can communicate with humans while performing various tasks in the real world. In the future, we will expand our approach with reinforcement learning to reduce the need for expert-defined action trajectories. Furthermore, a reinforcement learner may explore more dexterous object manipulation with diversified action trajectories. With more realistic action execution, we will attempt to tackle the problem of sim-to-real transfer. Lastly, diversifying our action repertoire will inevitably lead to more diverse natural language descriptions, which we can tackle by employing a pretrained Transformer-based large language model as a language encoder.

\begin{comment}
\section*{Acknowledgement(s)}

An unnumbered section, e.g.\ \verb"\section*{Acknowledgements}", may be used for thanks, etc.\ if required and included \emph{in the non-anonymous version} before any Notes or References.

\end{comment}
\section*{Disclosure statement}

The authors report there are no competing interests to declare.

\section*{Funding}

This work was supported by the German Research Foundation (DFG) under Project TRR 169 Crossmodal Learning (CML), LeCareBot, IDEAS, and MoReSpace.

\begin{comment}

\section*{Notes on contributor(s)}

An unnumbered section, e.g.\ \verb"\section*{Notes on contributors}", may be included \emph{in the non-anonymous version} if required. A photograph may be added if requested.

\section*{Nomenclature/Notation}

An unnumbered section, e.g.\ \verb"\section*{Nomenclature}" (or \verb"\section*{Notation}"), may be included if required, before any Notes or References.

\section*{Notes}

An unnumbered `Notes' section may be included before the References (if using the \verb"endnotes" package, use the command \verb"\theendnotes" where the notes are to appear, instead of creating a \verb"\section*").
\end{comment}

%\section{References}
\bibliographystyle{chicago2}%\\bibliographystyle{plainnat}
\bibliography{references}

\begin{thebibliography}{}

\bibitem[\protect\citeauthoryear{Abramson, Ahuja, Brussee, Carnevale, Cassin,
  Clark, Dudzik, Georgiev, Guy, Harley, Hill, Hung, Kenton, Landon, Lillicrap,
  Mathewson, Muldal, Santoro, Savinov, Varma, Wayne, Wong, Yan, and
  Zhu}{Abramson et~al.}{2020}]{abramson2020imitating}
Abramson, J., A.~Ahuja, A.~Brussee, F.~Carnevale, M.~Cassin, S.~Clark,
  A.~Dudzik, P.~Georgiev, A.~Guy, T.~Harley, F.~Hill, A.~Hung, Z.~Kenton,
  J.~Landon, T.~P. Lillicrap, K.~W. Mathewson, A.~Muldal, A.~Santoro,
  N.~Savinov, V.~Varma, G.~Wayne, N.~Wong, C.~Yan, and R.~Zhu.  2020.
\newblock Imitating interactive intelligence.
\newblock {\em arXiv preprint arXiv:2012.05672\/}.

\bibitem[\protect\citeauthoryear{Ahn, Brohan, Brown, Chebotar, Cortes, David,
  Finn, Fu, Gopalakrishnan, Hausman, Herzog, Ho, Hsu, Ibarz, Ichter, Irpan,
  Jang, Ruano, Jeffrey, Jesmonth, Joshi, Julian, Kalashnikov, Kuang, Lee,
  Levine, Lu, Luu, Parada, Pastor, Quiambao, Rao, Rettinghouse, Reyes,
  Sermanet, Sievers, Tan, Toshev, Vanhoucke, Xia, Xiao, Xu, Xu, Yan, and
  Zeng}{Ahn et~al.}{2022}]{saycan2022arxiv}
Ahn, M., A.~Brohan, N.~Brown, Y.~Chebotar, O.~Cortes, B.~David, C.~Finn, C.~Fu,
  K.~Gopalakrishnan, K.~Hausman, A.~Herzog, D.~Ho, J.~Hsu, J.~Ibarz, B.~Ichter,
  A.~Irpan, E.~Jang, R.~J. Ruano, K.~Jeffrey, S.~Jesmonth, N.~Joshi, R.~Julian,
  D.~Kalashnikov, Y.~Kuang, K.-H. Lee, S.~Levine, Y.~Lu, L.~Luu, C.~Parada,
  P.~Pastor, J.~Quiambao, K.~Rao, J.~Rettinghouse, D.~Reyes, P.~Sermanet,
  N.~Sievers, C.~Tan, A.~Toshev, V.~Vanhoucke, F.~Xia, T.~Xiao, P.~Xu, S.~Xu,
  M.~Yan, and A.~Zeng.  2022.
\newblock Do as {I} can and not as {I} say: Grounding language in robotic
  affordances.
\newblock {\em arXiv preprint arXiv:2204.01691\/}.

\bibitem[\protect\citeauthoryear{Antunes, Laflaquiere, Ogata, and
  Cangelosi}{Antunes et~al.}{2019}]{antunes2019multitimescale}
Antunes, A., A.~Laflaquiere, T.~Ogata, and A.~Cangelosi.  2019.
\newblock A bi-directional multiple timescales {LSTM} model for grounding of
  actions and verbs.
\newblock In {\em 2019 IEEE/RSJ International Conference on Intelligent Robots
  and Systems (IROS)}, pp.\  2614--2621.

\bibitem[\protect\citeauthoryear{Aravena, Hurtado, Riveros, Cardona, Manes, and
  Ib{\'a}{\~n}ez}{Aravena et~al.}{2010}]{aravena2010applauding}
Aravena, P., E.~Hurtado, R.~Riveros, J.~F. Cardona, F.~Manes, and
  A.~Ib{\'a}{\~n}ez.  2010.
\newblock Applauding with closed hands: neural signature of action-sentence
  compatibility effects.
\newblock {\em PloS ONE\/}~{\em 5\/}(7), e11751.

\bibitem[\protect\citeauthoryear{Arevalo, Solorio, Montes-y Gómez, and
  Gonz{\'a}lez}{Arevalo et~al.}{2020}]{arevalo2020gated}
Arevalo, J., T.~Solorio, M.~Montes-y Gómez, and F.~A. Gonz{\'a}lez.  2020.
\newblock Gated multimodal networks.
\newblock {\em Neural Computing and Applications\/}~{\em 32\/}(14),
  10209--10228.

\bibitem[\protect\citeauthoryear{Bisk, Holtzman, Thomason, Andreas, Bengio,
  Chai, Lapata, Lazaridou, May, Nisnevich, Pinto, and Turian}{Bisk
  et~al.}{2020}]{bisk-etal-2020-experience}
Bisk, Y., A.~Holtzman, J.~Thomason, J.~Andreas, Y.~Bengio, J.~Chai, M.~Lapata,
  A.~Lazaridou, J.~May, A.~Nisnevich, N.~Pinto, and J.~Turian.  2020, November.
\newblock Experience grounds language.
\newblock In {\em Proceedings of the 2020 Conference on Empirical Methods in
  Natural Language Processing}, pp.\  8718--8735. Association for Computational
  Linguistics.

\bibitem[\protect\citeauthoryear{Brown, Mann, Ryder, Subbiah, Kaplan, Dhariwal,
  Neelakantan, Shyam, Sastry, Askell, et~al.}{Brown
  et~al.}{2020}]{brown2020language}
Brown, T., B.~Mann, N.~Ryder, M.~Subbiah, J.~D. Kaplan, P.~Dhariwal,
  A.~Neelakantan, P.~Shyam, G.~Sastry, A.~Askell, et~al.  2020.
\newblock Language models are few-shot learners.
\newblock {\em Advances in Neural Information Processing Systems\/}~{\em 33},
  1877--1901.

\bibitem[\protect\citeauthoryear{Canals and Mor}{Canals and
  Mor}{2023}]{canals_mor_2023}
Canals, L. and Y.~Mor.  2023.
\newblock Towards a signature pedagogy for technology-enhanced task-based
  language teaching: Defining its design principles.
\newblock {\em ReCALL\/}~{\em 35\/}(1), 4–18.

\bibitem[\protect\citeauthoryear{Devlin, Chang, Lee, and Toutanova}{Devlin
  et~al.}{2019}]{devlin2019bert}
Devlin, J., M.-W. Chang, K.~Lee, and K.~Toutanova.  2019, June.
\newblock {BERT}: Pre-training of deep bidirectional transformers for language
  understanding.
\newblock In {\em Proceedings of the 2019 Conference of the North {A}merican
  Chapter of the Association for Computational Linguistics: Human Language
  Technologies, Volume 1 (Long and Short Papers)}, Minneapolis, Minnesota, pp.\
   4171--4186. Association for Computational Linguistics.

\bibitem[\protect\citeauthoryear{Eisermann, Lee, Weber, and Wermter}{Eisermann
  et~al.}{2021}]{ELWW21}
Eisermann, A., J.~H. Lee, C.~Weber, and S.~Wermter.  2021, Jul.
\newblock Generalization in multimodal language learning from simulation.
\newblock In {\em Proceedings of the International Joint Conference on Neural
  Networks (IJCNN 2021)}.

\bibitem[\protect\citeauthoryear{Glenberg and Kaschak}{Glenberg and
  Kaschak}{2002}]{glenberg2002grounding}
Glenberg, A.~M. and M.~P. Kaschak.  2002.
\newblock Grounding language in action.
\newblock {\em Psychonomic Bulletin \& Review\/}~{\em 9\/}(3), 558--565.

\bibitem[\protect\citeauthoryear{Hatori, Kikuchi, Kobayashi, Takahashi, Tsuboi,
  Unno, Ko, and Tan}{Hatori et~al.}{2018}]{hatori2018interactively}
Hatori, J., Y.~Kikuchi, S.~Kobayashi, K.~Takahashi, Y.~Tsuboi, Y.~Unno, W.~Ko,
  and J.~Tan.  2018.
\newblock Interactively picking real-world objects with unconstrained spoken
  language instructions.
\newblock In {\em 2018 IEEE International Conference on Robotics and Automation
  (ICRA)}, pp.\  3774--3781. IEEE.

\bibitem[\protect\citeauthoryear{Hauk, Johnsrude, and Pulverm{\"u}ller}{Hauk
  et~al.}{2004}]{hauk2004somatotopic}
Hauk, O., I.~Johnsrude, and F.~Pulverm{\"u}ller.  2004.
\newblock Somatotopic representation of action words in human motor and
  premotor cortex.
\newblock {\em Neuron\/}~{\em 41\/}(2), 301--307.

\bibitem[\protect\citeauthoryear{Heinrich, Yao, Hinz, Liu, Hummel, Kerzel,
  Weber, and Wermter}{Heinrich et~al.}{2020}]{heinrich2020}
Heinrich, S., Y.~Yao, T.~Hinz, Z.~Liu, T.~Hummel, M.~Kerzel, C.~Weber, and
  S.~Wermter.  2020.
\newblock Crossmodal language grounding in an embodied neurocognitive model.
\newblock {\em Frontiers in Neurorobotics\/}~{\em 14}, 52.

\bibitem[\protect\citeauthoryear{Hochreiter and Schmidhuber}{Hochreiter and
  Schmidhuber}{1997}]{hochreiter1997long}
Hochreiter, S. and J.~Schmidhuber.  1997.
\newblock Long short-term memory.
\newblock {\em Neural Computation\/}~{\em 9\/}(8), 1735--1780.

\bibitem[\protect\citeauthoryear{Irshad, Ma, and Kira}{Irshad
  et~al.}{2021}]{irshadhcm2021}
Irshad, M.~Z., C.-Y. Ma, and Z.~Kira.  2021.
\newblock Hierarchical cross-modal agent for robotics vision-and-language
  navigation.
\newblock In {\em 2021 IEEE International Conference on Robotics and Automation
  (ICRA)}, pp.\  13238--13246.

\bibitem[\protect\citeauthoryear{Jaegle, Borgeaud, Alayrac, Doersch, Ionescu,
  Ding, Koppula, Zoran, Brock, Shelhamer, et~al.}{Jaegle
  et~al.}{2021}]{perceiverio}
Jaegle, A., S.~Borgeaud, J.-B. Alayrac, C.~Doersch, C.~Ionescu, D.~Ding,
  S.~Koppula, D.~Zoran, A.~Brock, E.~Shelhamer, et~al.  2021.
\newblock Perceiver io: A general architecture for structured inputs \&
  outputs.
\newblock In {\em International Conference on Learning Representations}.

\bibitem[\protect\citeauthoryear{Jang, Irpan, Khansari, Kappler, Ebert, Lynch,
  Levine, and Finn}{Jang et~al.}{2021}]{bc-z}
Jang, E., A.~Irpan, M.~Khansari, D.~Kappler, F.~Ebert, C.~Lynch, S.~Levine, and
  C.~Finn.  2021.
\newblock {BC}-z: Zero-shot task generalization with robotic imitation
  learning.
\newblock In {\em 5th Annual Conference on Robot Learning}.

\bibitem[\protect\citeauthoryear{Jiang, Gupta, Zhang, Wang, Dou, Chen, Fei-Fei,
  Anandkumar, Zhu, and Fan}{Jiang et~al.}{2022}]{vima2022}
Jiang, Y., A.~Gupta, Z.~Zhang, G.~Wang, Y.~Dou, Y.~Chen, L.~Fei-Fei,
  A.~Anandkumar, Y.~Zhu, and L.~Fan.  2022.
\newblock {VIMA}: General robot manipulation with multimodal prompts.

\bibitem[\protect\citeauthoryear{Kaschak, Madden, Therriault, Yaxley, Aveyard,
  Blanchard, and Zwaan}{Kaschak et~al.}{2005}]{kaschak2005perception}
Kaschak, M.~P., C.~J. Madden, D.~J. Therriault, R.~H. Yaxley, M.~Aveyard, A.~A.
  Blanchard, and R.~A. Zwaan.  2005.
\newblock Perception of motion affects language processing.
\newblock {\em Cognition\/}~{\em 94\/}(3), B79--B89.

\bibitem[\protect\citeauthoryear{Kerzel, Pekarek-Rosin, Strahl, Heinrich, and
  Wermter}{Kerzel et~al.}{2020}]{kerzel2020teaching}
Kerzel, M., T.~Pekarek-Rosin, E.~Strahl, S.~Heinrich, and S.~Wermter.  2020.
\newblock Teaching {NICO} how to grasp: an empirical study on crossmodal social
  interaction as a key factor for robots learning from humans.
\newblock {\em Frontiers in Neurorobotics\/}~{\em 14}, 28.

\bibitem[\protect\citeauthoryear{Kerzel, Strahl, Magg, Navarro-Guerrero,
  Heinrich, and Wermter}{Kerzel et~al.}{2017}]{kerzel2017nico}
Kerzel, M., E.~Strahl, S.~Magg, N.~Navarro-Guerrero, S.~Heinrich, and
  S.~Wermter.  2017.
\newblock {NICO}—{N}euro-{I}nspired {CO}mpanion: A developmental humanoid
  robot platform for multimodal interaction.
\newblock In {\em 2017 26th IEEE International Symposium on Robot and Human
  Interactive Communication (RO-MAN)}, pp.\  113--120. IEEE.

\bibitem[\protect\citeauthoryear{Kingma and Ba}{Kingma and
  Ba}{2015}]{kingma2015adam}
Kingma, D.~P. and J.~Ba.  2015.
\newblock Adam: {A} method for stochastic optimization.
\newblock In {\em 3rd International Conference on Learning Representations,
  {ICLR}, San Diego, CA, USA, May 7-9}.

\bibitem[\protect\citeauthoryear{Lu, Batra, Parikh, and Lee}{Lu
  et~al.}{2019}]{lu2019vilbert}
Lu, J., D.~Batra, D.~Parikh, and S.~Lee.  2019.
\newblock {ViLBERT}: Pretraining task-agnostic visiolinguistic representations
  for vision-and-language tasks.
\newblock In H.~Wallach, H.~Larochelle, A.~Beygelzimer, F.~d\textquotesingle
  Alch\'{e}-Buc, E.~Fox, and R.~Garnett (Eds.), {\em Advances in Neural
  Information Processing Systems}, Volume~32. Curran Associates, Inc.

\bibitem[\protect\citeauthoryear{Lynch and Sermanet}{Lynch and
  Sermanet}{2021}]{lynch2021language}
Lynch, C. and P.~Sermanet.  2021.
\newblock Language conditioned imitation learning over unstructured data.
\newblock In D.~A. Shell, M.~Toussaint, and M.~A. Hsieh (Eds.), {\em Robotics:
  Science and System XVII}.

\bibitem[\protect\citeauthoryear{Meteyard, Bahrami, and Vigliocco}{Meteyard
  et~al.}{2007}]{meteyard2007motion}
Meteyard, L., B.~Bahrami, and G.~Vigliocco.  2007.
\newblock Motion detection and motion verbs: Language affects low-level visual
  perception.
\newblock {\em Psychological Science\/}~{\em 18\/}(11), 1007--1013.
\newblock PMID: 17958716.

\bibitem[\protect\citeauthoryear{Ogata, Murase, Tani, Komatani, and
  Okuno}{Ogata et~al.}{2007}]{ogata2007parametricbias}
Ogata, T., M.~Murase, J.~Tani, K.~Komatani, and H.~G. Okuno.  2007.
\newblock Two-way translation of compound sentences and arm motions by
  recurrent neural networks.
\newblock In {\em 2007 IEEE/RSJ International Conference on Intelligent Robots
  and Systems}, pp.\  1858--1863.

\bibitem[\protect\citeauthoryear{{\"O}zdemir, Kerzel, Weber, Lee, and
  Wermter}{{\"O}zdemir et~al.}{2022}]{pgae2022}
{\"O}zdemir, O., M.~Kerzel, C.~Weber, J.~H. Lee, and S.~Wermter.  2022.
\newblock Learning flexible translation between robot actions and language
  descriptions.
\newblock In E.~Pimenidis, P.~Angelov, C.~Jayne, A.~Papaleonidas, and M.~Aydin
  (Eds.), {\em Artificial Neural Networks and Machine Learning -- ICANN 2022},
  Cham, pp.\  246--257. Springer Nature Switzerland.

\bibitem[\protect\citeauthoryear{{\"O}zdemir, Kerzel, and Wermter}{{\"O}zdemir
  et~al.}{2021}]{Oezdemir_2021_ICDL}
{\"O}zdemir, O., M.~Kerzel, and S.~Wermter.  2021, Aug.
\newblock Embodied language learning with paired variational autoencoders.
\newblock In {\em 2021 IEEE International Conference on Development and
  Learning (ICDL)}, pp.\  1--6.

\bibitem[\protect\citeauthoryear{Radford, Kim, Hallacy, Ramesh, Goh, Agarwal,
  Sastry, Askell, Mishkin, Clark, et~al.}{Radford
  et~al.}{2021}]{radford2021clip}
Radford, A., J.~W. Kim, C.~Hallacy, A.~Ramesh, G.~Goh, S.~Agarwal, G.~Sastry,
  A.~Askell, P.~Mishkin, J.~Clark, et~al.  2021.
\newblock Learning transferable visual models from natural language
  supervision.
\newblock In {\em International Conference on Machine Learning}, pp.\
  8748--8763. PMLR.

\bibitem[\protect\citeauthoryear{Radford, Wu, Child, Luan, Amodei, and
  Sutskever}{Radford et~al.}{2019}]{radford2019language}
Radford, A., J.~Wu, R.~Child, D.~Luan, D.~Amodei, and I.~Sutskever.  2019.
\newblock Language models are unsupervised multitask learners.

\bibitem[\protect\citeauthoryear{Raffel, Shazeer, Roberts, Lee, Narang, Matena,
  Zhou, Li, and Liu}{Raffel et~al.}{2020}]{2020raffelt5}
Raffel, C., N.~Shazeer, A.~Roberts, K.~Lee, S.~Narang, M.~Matena, Y.~Zhou,
  W.~Li, and P.~J. Liu.  2020.
\newblock Exploring the limits of transfer learning with a unified text-to-text
  transformer.
\newblock {\em Journal of Machine Learning Research\/}~{\em 21\/}(140), 1--67.

\bibitem[\protect\citeauthoryear{Reed, Zolna, Parisotto, Colmenarejo, Novikov,
  Barth-Maron, Gimenez, Sulsky, Kay, Springenberg, Eccles, Bruce, Razavi,
  Edwards, Heess, Chen, Hadsell, Vinyals, Bordbar, and de~Freitas}{Reed
  et~al.}{2022}]{gato2022}
Reed, S., K.~Zolna, E.~Parisotto, S.~G. Colmenarejo, A.~Novikov,
  G.~Barth-Maron, M.~Gimenez, Y.~Sulsky, J.~Kay, J.~T. Springenberg, T.~Eccles,
  J.~Bruce, A.~Razavi, A.~Edwards, N.~Heess, Y.~Chen, R.~Hadsell, O.~Vinyals,
  M.~Bordbar, and N.~de~Freitas.  2022.
\newblock A generalist agent.

\bibitem[\protect\citeauthoryear{Shao, Migimatsu, Zhang, Yang, and Bohg}{Shao
  et~al.}{2020}]{shao2020concept2robot}
Shao, L., T.~Migimatsu, Q.~Zhang, K.~Yang, and J.~Bohg.  2020.
\newblock Concept2robot: Learning manipulation concepts from instructions and
  human demonstrations.
\newblock In {\em Proceedings of Robotics: Science and Systems (RSS)}.

\bibitem[\protect\citeauthoryear{Shridhar, Manuelli, and Fox}{Shridhar
  et~al.}{2021}]{shridhar2021cliport}
Shridhar, M., L.~Manuelli, and D.~Fox.  2021.
\newblock {CLIPort}: What and where pathways for robotic manipulation.
\newblock In {\em Proceedings of the 5th Conference on Robot Learning (CoRL)}.

\bibitem[\protect\citeauthoryear{Shridhar, Manuelli, and Fox}{Shridhar
  et~al.}{2022}]{shridhar2022peract}
Shridhar, M., L.~Manuelli, and D.~Fox.  2022.
\newblock {Perceiver-Actor}: A multi-task transformer for robotic manipulation.
\newblock In {\em Proceedings of the 6th Conference on Robot Learning (CoRL)}.

\bibitem[\protect\citeauthoryear{Shridhar, Mittal, and Hsu}{Shridhar
  et~al.}{2020}]{shridhar2018interactive}
Shridhar, M., D.~Mittal, and D.~Hsu.  2020.
\newblock {INGRESS}: Interactive visual grounding of referring expressions.
\newblock {\em The International Journal of Robotics Research\/}~{\em
  39\/}(2-3), 217--232.

\bibitem[\protect\citeauthoryear{van Elk, van Schie, Zwaan, and Bekkering}{van
  Elk et~al.}{2010}]{van2010functional}
van Elk, M., H.~T. van Schie, R.~A. Zwaan, and H.~Bekkering.  2010.
\newblock The functional role of motor activation in language processing: Motor
  cortical oscillations support lexical-semantic retrieval.
\newblock {\em Neuroimage\/}~{\em 50\/}(2), 665--677.

\bibitem[\protect\citeauthoryear{Vaswani, Shazeer, Parmar, Uszkoreit, Jones,
  Gomez, Kaiser, and Polosukhin}{Vaswani et~al.}{2017}]{vaswani2017attention}
Vaswani, A., N.~Shazeer, N.~Parmar, J.~Uszkoreit, L.~Jones, A.~N. Gomez,
  {\L}.~Kaiser, and I.~Polosukhin.  2017.
\newblock Attention is all you need.
\newblock In {\em Advances in Neural Information Processing Systems}, pp.\
  5998--6008.

\bibitem[\protect\citeauthoryear{Winter, Dudschig, Miller, Ulrich, and
  Kaup}{Winter et~al.}{2022}]{winter2022action}
Winter, A., C.~Dudschig, J.~Miller, R.~Ulrich, and B.~Kaup.  2022.
\newblock The action-sentence compatibility effect (ace): Meta-analysis of a
  benchmark finding for embodiment.
\newblock {\em Acta Psychologica\/}~{\em 230}, 103712.

\bibitem[\protect\citeauthoryear{Yamada, Matsunaga, and Ogata}{Yamada
  et~al.}{2018}]{yamada2018paired}
Yamada, T., H.~Matsunaga, and T.~Ogata.  2018.
\newblock Paired recurrent autoencoders for bidirectional translation between
  robot actions and linguistic descriptions.
\newblock {\em IEEE Robotics and Automation Letters\/}~{\em 3\/}(4),
  3441--3448.

\bibitem[\protect\citeauthoryear{Zeng, Florence, Tompson, Welker, Chien,
  Attarian, Armstrong, Krasin, Duong, Sindhwani, and Lee}{Zeng
  et~al.}{2020}]{zeng2020transporter}
Zeng, A., P.~Florence, J.~Tompson, S.~Welker, J.~Chien, M.~Attarian,
  T.~Armstrong, I.~Krasin, D.~Duong, V.~Sindhwani, and J.~Lee.  2020.
\newblock Transporter networks: Rearranging the visual world for robotic
  manipulation.
\newblock In J.~Kober, F.~Ramos, and C.~J. Tomlin (Eds.), {\em 4th Conference
  on Robot Learning, CoRL 2020, 16-18 November 2020, Virtual Event / Cambridge,
  MA, {USA}}, Volume 155 of {\em Proceedings of Machine Learning Research},
  pp.\  726--747. {PMLR}.

\end{thebibliography}
%\begin{thebibliography}{}

%\end{thebibliography}
\bigskip

\end{document}